\documentclass[11pt]{article}
\usepackage{histocrypt}
\usepackage{mathptmx}
\usepackage{url}
\usepackage{latexsym}
\usepackage{graphicx}
\usepackage{subcaption}
\usepackage{amsmath}
\usepackage{amssymb}

\usepackage{float}
\usepackage{placeins}

\newlength{\imgwidth}
\newcommand{\customfig}[4]{
  \begin{figure}[#4]
    \centering
    \setlength{\fboxsep}{1pt} 
    \setlength{\fboxrule}{1pt} 
    \addtolength{\imgwidth}{-2\fboxsep}
    \addtolength{\imgwidth}{-2\fboxrule}
    \fbox{
      \includegraphics[width=\imgwidth]{#1}
    }
    \caption{#2}
    \label{#3}
  \end{figure}
}

\usepackage[framemethod=TikZ]{mdframed}

 \usepackage{xcolor}

\usepackage{tabularx}

\usepackage{enumitem}

\title{Subtle Signs of Scribal Intent in the Voynich Manuscript}

 \author{Andrew Steckley, PhD \\
   QuantumLynx Research \\
   {\tt andrew@quantumlynxresearch.com}\\\And
   Noah Steckley \\
   QuantumLynx Research \\
   {\tt noah@quantumlynxresearch.com} 
   \\}

\begin{document}
\maketitle

\begin{abstract}
This study explores the cryptic Voynich Manuscript, by looking for subtle signs of scribal intent hidden in overlooked features of the “Voynichese” script. The findings indicate that distributions of tokens within paragraphs vary significantly based on positions defined not only by elements intrinsic to the script such as paragraph and line boundaries but also by extrinsic elements, namely the hand-drawn illustrations of plants.
\end{abstract}

\section{Introduction}

The Voynich Manuscript, with its inscrutable script and peculiar illustrations, has been extensively studied by amateur and professional researchers for almost a century. Despite this, there is no consensus as to its origin, authorship, or the meaning of its unusual script. There is even disagreement as to whether it has meaning at all.

Many reasonable arguments have been put forth supporting divergent scenarios: either that it conveys meaningful content or that it is meaningless, crafted only for visual appearance. Most researchers who have studied the script have implicitly assumed that it contains meaning, focusing their efforts on finding grammatical structures or statistical signatures that would identify the most likely known language of origin from which Voynichese may have been derived or encoded. Those efforts that have aimed objectively at the more primary question —whether it does or does not have meaning in the first place— have predominantly looked for evidence of the script sharing various statistical properties with known languages, the implication being that it is therefore not just a meaningless imitation.

A few studies have explored the possibility of the script being meaningless. Rugg demonstrated a technique, inspired by a Cardan table and grille cipher, that could explain the script’s general appearance and feasibly produce a manuscript of similar size manually, within a reasonable timeframe \cite{rugg_elegant_2004}. Zandbergen further analyzed the technique to show how it could produce some of the apparent statistical structure observed in the Voynichese script \cite{zandbergen_cardan_2021}. Neither researcher sought, nor claimed, to explain all of the observed structure, but they did show that a script produced using a language simulation device, although meaningless, could still exhibit some linguistic structure as an artifact of the simulation process. Gaskell and Bowern showed that samples of “gibberish” script also shared certain statistical properties with the Voynichese script \cite{gaskell_gibberish_2022}. The majority opinion, however, remains that the Voynichese script has meaning and will eventually be deciphered.

This broad-brush summary describes not only the historic published research, but also the many blogs and forum discussions within the community of Voynich enthusiasts that has been growing continually since the Internet made images and details of the Voynich Manuscript more widely available. That community includes a large number of amateur researchers who have bought wholesale into the assumption of meaning,  believing they have found concrete connections to known languages or that they have even deciphered specific words or sections of the manuscript. None of these claims,  however, have withstood even modest levels of critical review.

It should be acknowledged that an unexpected solution could still emerge from the online community; it would be shortsighted to overlook the role of guesswork and intuition in scientific discoveries. Nevertheless, advances in understanding the Voynichese script appear to have plateaued over the past decade, despite increased scholarly involvement and public interest, and it seems clear that fresh perspectives in attacking the problem are needed.

We suggest that a paramount objective ought to be the determination of whether the script contains meaningful content that can be deciphered or whether it presents merely the appearance of such content. A determination, irrespective of its outcome, would significantly sharpen and enhance the effectiveness of ongoing research efforts.

Toward these ends, our research seeks to uncover evidence of the scribe’s intentions; the supposition is that subtle signs may be found in overlooked patterns in the script. In the present study, we focus on the statistical distribution of “word” tokens and their placement relative to structural features like line beginnings and endings, and adjacency to intricate plant drawings that disrupt the script's flow.

In the following section, we describe further how such statistical evidence might indicate the scribe's purpose. Following that, in Sections 3, 4, and 5, we describe how the transliterated Voynichese data were prepared for use in this study. Sections 6 and 7 then describe the two major analyses performed and discuss their results. Section 8 summarizes the conclusions drawn directly from the analytical results. Finally, in Section 9, we discuss the implications of these conclusions, particularly with respect to whether the scribe intended the Voynichese script to convey meaningful or meaningless content.

\section{An Overlooked Feature of the Manuscript}

One prominent feature of the manuscript is that the lines of Voynichese script appear to have been intentionally written out to visually conform to the outlines of previously drawn illustrations. In most cases, when encountering the intrusion of a drawing, the scribe has skipped over it to resume writing on the far side of the drawing, or within space between drawing elements when it is of  sufficient size. Various researchers have noted this contouring of the script's margins, but it has received minimal investigative effort.

Figure \ref{fig:sample_script_conformities} shows an example of this shape conformity, which is consistently seen across all folio pages featuring large drawings in the manuscript. This phenomenon prompts an intriguing question: Did the scribe deliberately select tokens of specific lengths to create this visual effect potentially altering —or simply without regard to — meaningful content in the process?

Gaskell and Bowern looked briefly at whether student volunteers, instructed to compose meaningless text wrapped around plant drawings in this way, were inclined to self-select the lengths of their made-up words to better achieve the desired result  \cite{gaskell_gibberish_2022}.  Their assessment, although worth reporting, was not surprising; it suggested that this feature of the manuscript may be common to gibberish documents. Beyond this, no other analytical efforts have focused on this visual aspect of the script.

Whether the tokens represent words in the manner of common written languages, or by some other structured means of converting thoughts into written script, it would seem to be a difficult task to achieve this level of visual conformity while preserving prescribed semantic content. We use the term “prescribed” here to refer to the idea that the meaning of the text is established before pen meets parchment to form each word. This content could originate from the thoughts of the scribe himself as he formulates words during the writing process, or it might involve the direct transcription of someone else’s words.

In the first case, he must frequently reconcile the word choices he would naturally make with what may be accommodated by the space remaining before a drawing.  In doing so, he must then continually adjust for any drifts in the prescribed meaning introduced by his contrived word choices. This is all possible, but at a cognitive cost. Achieving both the visual and semantic objectives is even more challenging if the scribe is an amanuensis, tasked with faithfully transcribing words from another document or taking dictation from another person.

On the other hand, if the scribe’s aims are to produce a merely visual result, while attempting only to simulate script appearing to have meaning and structure, then he is not constrained by prescribed content; he is free to choose words of suitable length at each opportunity.

\customfig{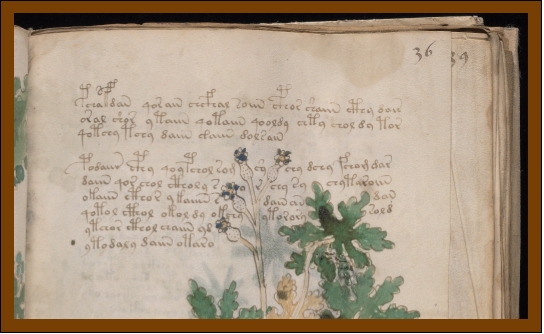}{Voynichese Script on Folio f36r. The script exhibits a noticeable conformity to the outline shapes of the drawn illustrations.}{fig:sample_script_conformities}{htbp}

Some researchers have noticed and analyzed the positional aspect of word tokens with respect to the beginning and ending of lines \cite{bunn_word_2022,bowern_linguistics_2021}. The present study, however, looks in detail at token positions not only at line boundaries, but adjacent to the hand-drawn illustrations.  One would not expect the drawings to be coupled to the syntactic content of the script, as might be conceivable in the case of tokens at the beginning and end of lines.\footnote{More conceivable would be an impact on the spacing between tokens and anomalies in the hand-written width of the glyphs. A separate study analyzing these features is in progress.}

However, this study found distinct and statistically significant differences in the populations of tokens that appear immediately before and after the intrusion of the drawings, as well as at the beginning and ending of lines.

\section{Transliterations}

Several transliterations of the Voynichese script have been compiled by various researchers as far back as the 1940s using various alphabets designed to represent the Voynichese glyphs. Zandbergen provides an excellent history and description of these transliterations \cite{zandbergen_text_2023}. He also designed a standard format called Intermediate Voynich Transliteration File Format (IVTFF) and has made several of the most comprehensive and reliable transliterations available in this format on his website. These standardized transliterations are invaluable for performing detailed analyses.  

For this study, we have used  the “Zandbergen-Landini” transliteration,\footnote{https:voynich.nu/data/ZL3a-n.txt} which is described and available on Zandbergen’s website. This transliteration contains comprehensive specifications of the location of the Voynichese elements, as well as indications of uncertain glyph identifications and uncertain token delimiting spaces.  It also indicates when two tokens within the same apparent line of text are separated by a portion of a drawn illustrations.

\section{Study Corpus}

It is apparent that the manuscript contains several different sections differing in layout style and illustration type.  Several researchers \cite{newbold_cipher_1928,dimperio_voynich_1978,zandbergen_analysis_2022} have grouped the manuscript’s folio pages according to these features and it is generally assumed that each group deals with a different topic.

In addition, it is believed that several individual scribes were involved in the creation of the manuscript. Using digital paleographic techniques, five separate scribes and the particular folios written by each have been proposed \cite{davis_how_2020}.

\customfig{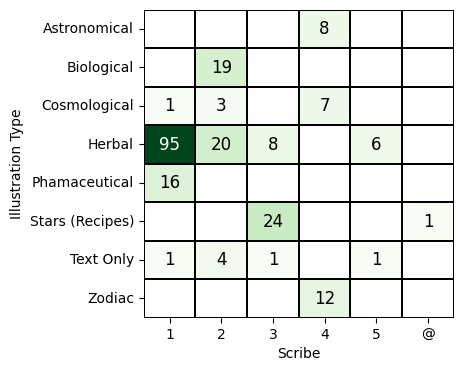}{Folio Page Count by Scribe and Illustration Type}{fig:folio_counts}{htbp}

To reduce variations in vocabulary that may result from the different topics being covered, and from variations in the mannerisms and styles of different scribes, we wanted to limit the analysis to a single topic and a single scribe.  Our reasoning for this is as follows: any analysis of a corpus consisting of heterogeneous folios increases the likelihood of significant attributes being obscured by the noise of multiple contributory factors. On the other hand, any conclusion once drawn from a more homogeneous corpus (e.g. that of a single scribe dealing with a single topic) may subsequently be given independent and explicit consideration as to whether it likely applies to the rest of the manuscript (e.g. other scribes dealing with other topics). 

The set of 95 folios of “Herbal” sections attributed to Davis’ “Scribe 1” was selected. As seen in Figure \ref{fig:folio_counts}, this is the largest homogeneous collection of folio pages. There are other topics containing more tokens per page, but this selection also contains the majority of drawings with the conforming script feature described earlier. 

We also wanted to restrict our analysis to the ‘paragraphs’ wherein a concept of position could be well defined, and to omit tokens that might only contribute statistical noise. 

These considerations resulted in the following criteria, which were applied to the full Landini-Zandbergen transliteration to produce the target corpus for this study.

\vspace{0.5\baselineskip}
\noindent \textbf{Include:}
\begin{itemize}[noitemsep,topsep=0pt,partopsep=0pt, leftmargin=*]
    \item Folios with ‘Herbal’ illustrations
    \item Folios penned by ‘Scribe 1’
    \item Lines of the script that were in ‘paragraph’ form (as opposed to labels or floating phrases)
\end{itemize}

\vspace{0.5\baselineskip}
\noindent \textbf{Exclude:}
\begin{itemize}[noitemsep,topsep=0pt,partopsep=0pt, leftmargin=*]
    \item Final token in each paragraph
    \item Any token that is ambiguous in the sense that the transliteration indicates more than one possibility for any of its glyphs 
    \item Any tokens where there is uncertainty as to whether the space before or after it is meant to delimit it from an adjacent token
\end{itemize}
\vspace{0.5\baselineskip}

Applying these criteria substantially reduced the data from the original transliteration — by roughly 80\% —  but this sacrifice of quantity for quality ensured that the study corpus\footnote{From the 39,020 tokens arranged in 5,389 lines on 227 folio pages, the study corpus extracted 7,660 tokens arranged in 1,223 lines on 95 folio pages.} was comprised of tokens that were all well defined  and unambiguous.

\section{Token Cohorts for Analysis}

Several sets of tokens were compiled from the study corpus, in order to provide separate cohorts for analytical comparisons.  These are summarized in Table \ref{tab:summary_of_cohorts} by the reference label used for each cohort throughout this paper.

\begin{table*}[htbp]
  \centering
  \includegraphics[width=\linewidth]{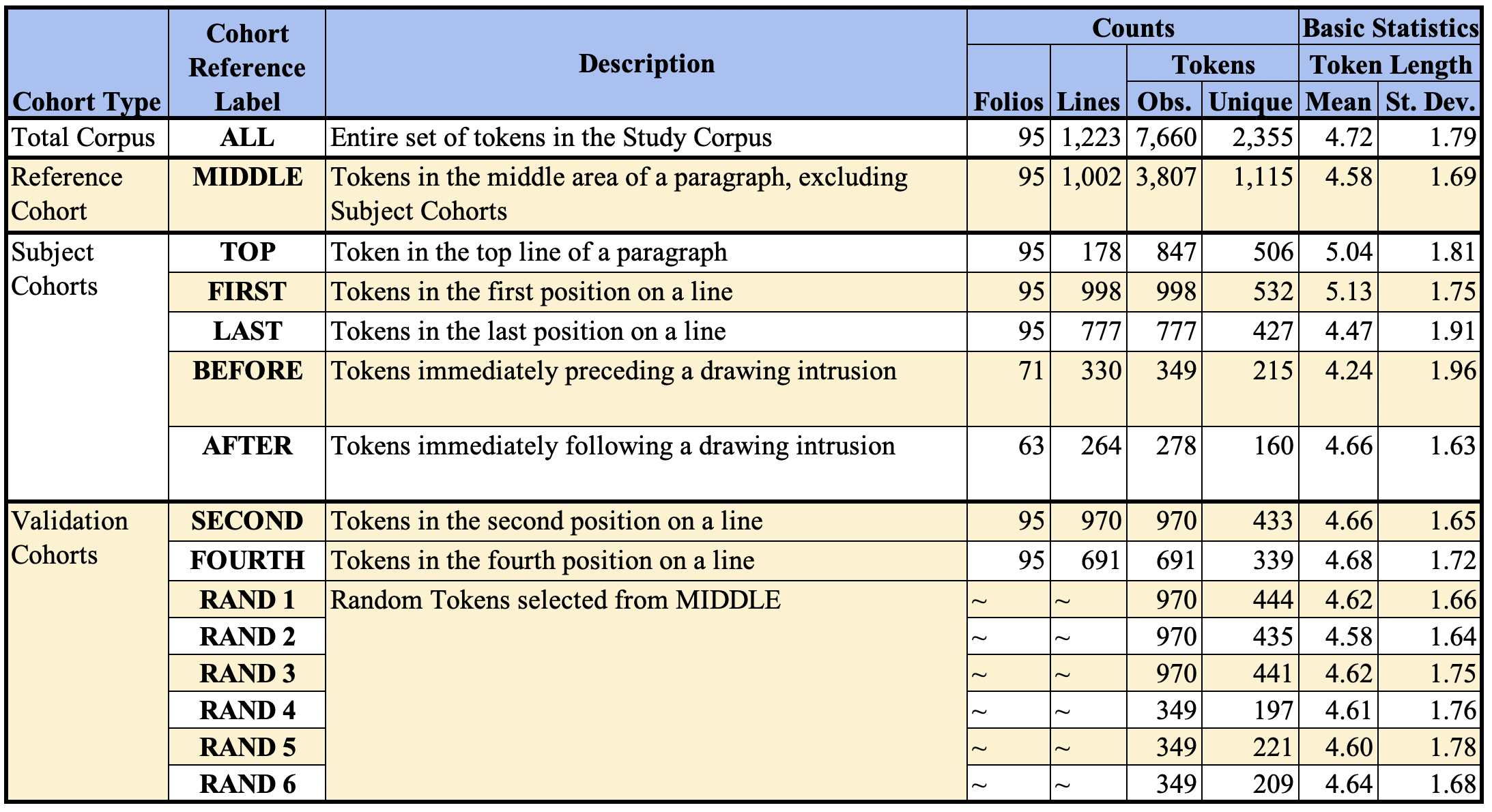}
  \caption{Summary of Cohorts}
  \label{tab:summary_of_cohorts}
\end{table*}

The reference cohort (MIDDLE) and the subject cohorts (TOP, FIRST, LAST, BEFORE, and AFTER) provided the main targets for the analyses. Mutual exclusivity was enforced on these cohorts, meaning that any token that would otherwise be in more than one of these cohorts was excluded from all of them. Two supplementary positional cohorts (SECOND and FOURTH), along with several randomized cohorts (RAND 1 through RAND 6), were used for additional validations. The  random cohorts correspond in size to the smaller subject cohorts, and were formed by drawing random selections of tokens from the MIDDLE cohort.

Note that any transliteration uncertainty of spacing to the left of a token on a line results in a range of uncertainty of its ordinal position. The net effect of this on these data is that 8.5\% of the tokens designated as being in the second position, and 20.8\% of those in the fourth position, may in fact belong in a higher position. It is for this reason that the SECOND and FOURTH cohorts were excluded from the set of subject cohorts, thereby ensuring that none of the main subject cohorts were affected by such spacing uncertainties. 

\customfig{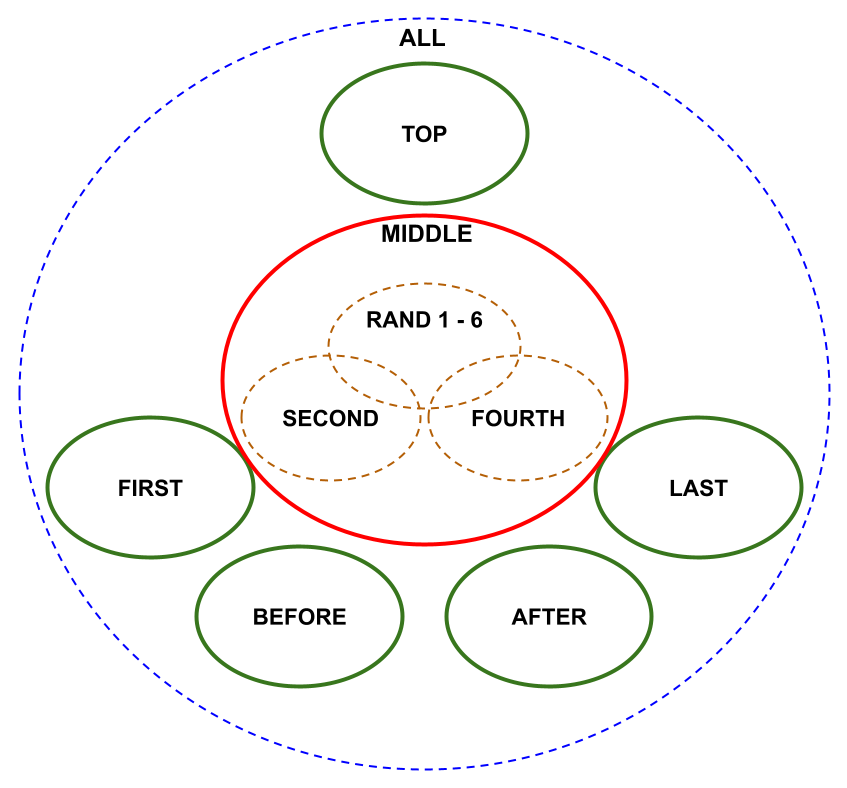}{Venn Diagram of Cohorts. \textit{Note that the reference and subject cohorts (shown with solid borders) are all mutually exclusive.}}{fig:venn_diagram}{H}

\begin{figure}[h]
  \centering

  \setlength{\fboxsep}{0pt} 
  \setlength{\fboxrule}{1pt} 

  \begin{minipage}{\linewidth}
    \fbox{ 
      \begin{minipage}{\linewidth}
        \newcommand{\subfigwidth}{0.48\linewidth}

        \begin{subfigure}[b]{\subfigwidth}
          \includegraphics[width=\linewidth]{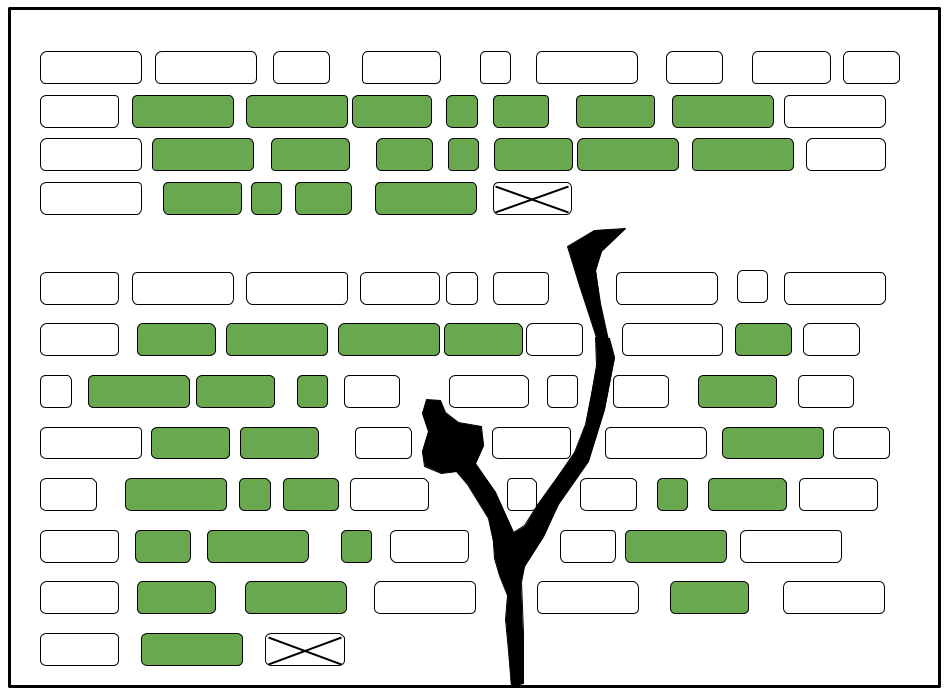}
          \caption{MIDDLE}
          \label{fig:schem_middle}
        \end{subfigure}
        \begin{subfigure}[b]{\subfigwidth}
          \includegraphics[width=\linewidth]{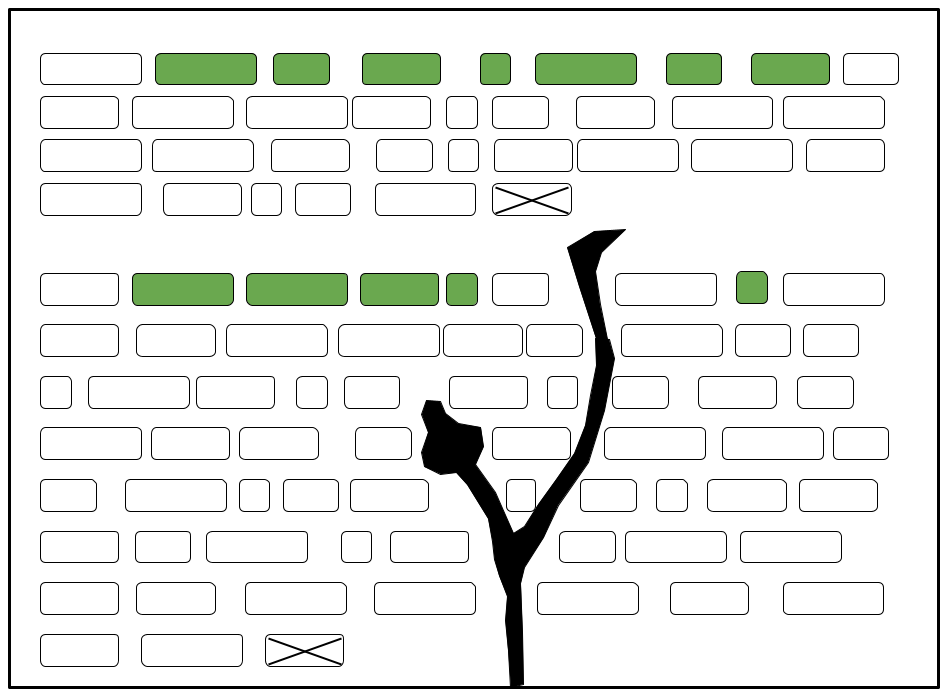}
          \caption{TOP}
          \label{fig:chem_top}
        \end{subfigure}

        \begin{subfigure}[b]{\subfigwidth}
          \includegraphics[width=\linewidth]{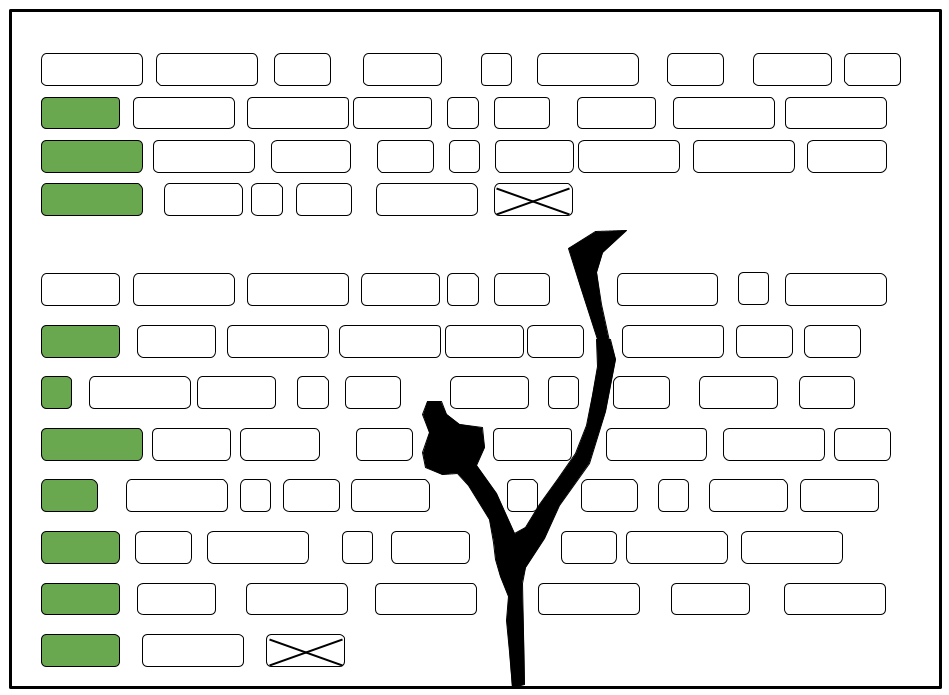}
          \caption{FIRST}
          \label{fig:schem_first}
        \end{subfigure}
        \begin{subfigure}[b]{\subfigwidth}
          \includegraphics[width=\linewidth]{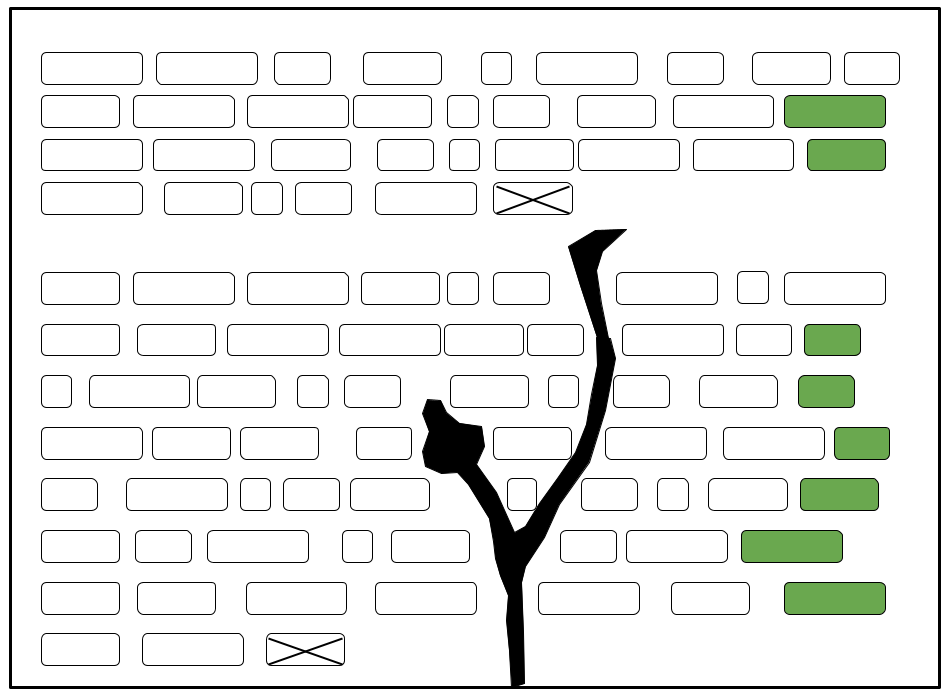}
          \caption{LAST}
          \label{fig:schem_last}
        \end{subfigure}

        \begin{subfigure}[b]{\subfigwidth}
          \includegraphics[width=\linewidth]{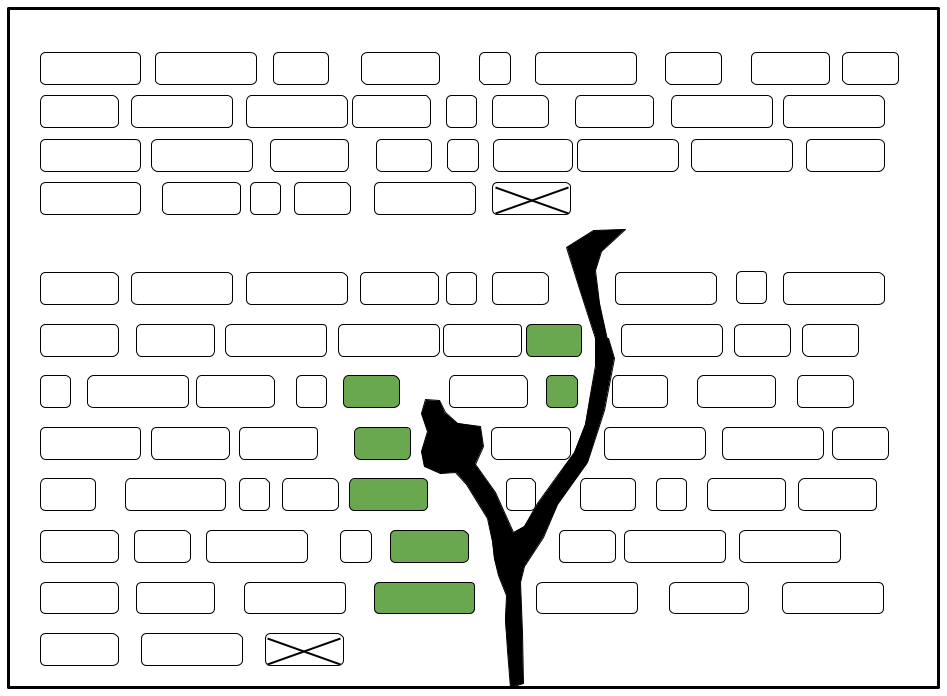}
          \caption{BEFORE}
          \label{fig:chem_pre}
        \end{subfigure}
        \begin{subfigure}[b]{\subfigwidth}
          \includegraphics[width=\linewidth]{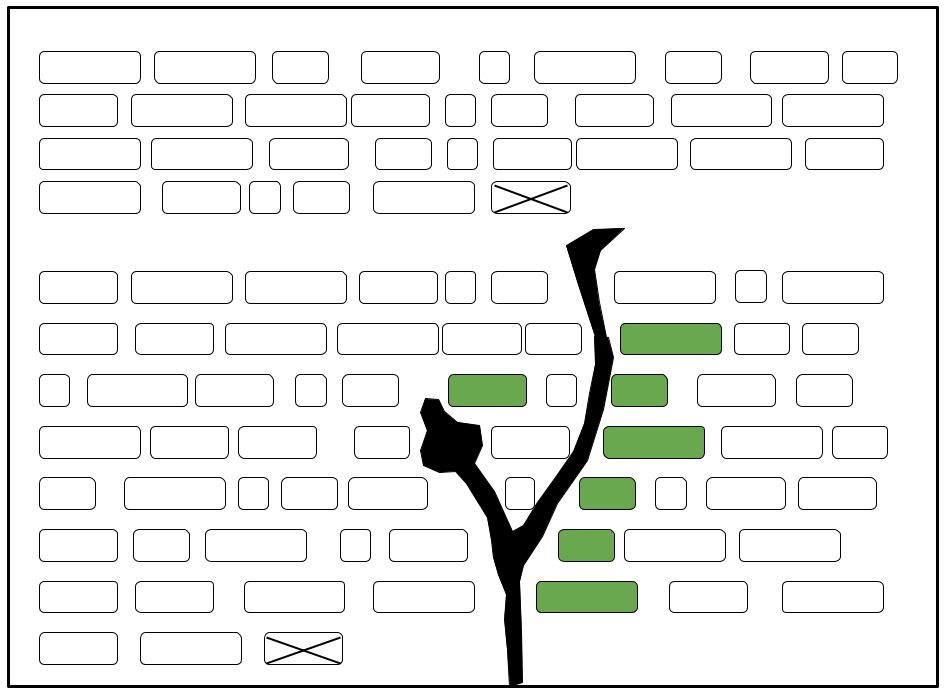}
          \caption{AFTER}
          \label{fig:schem_post}
        \end{subfigure}

        \begin{subfigure}[b]{\subfigwidth}
          \includegraphics[width=\linewidth]{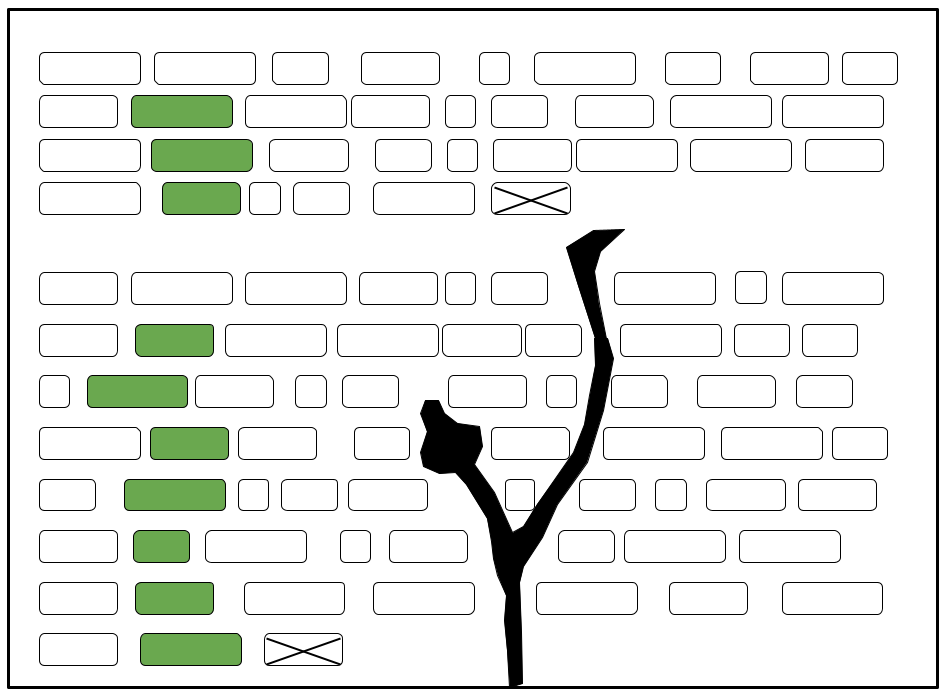}
          \caption{SECOND}
          \label{fig:schem_second}
        \end{subfigure}
        \begin{subfigure}[b]{\subfigwidth}
          \includegraphics[width=\linewidth]{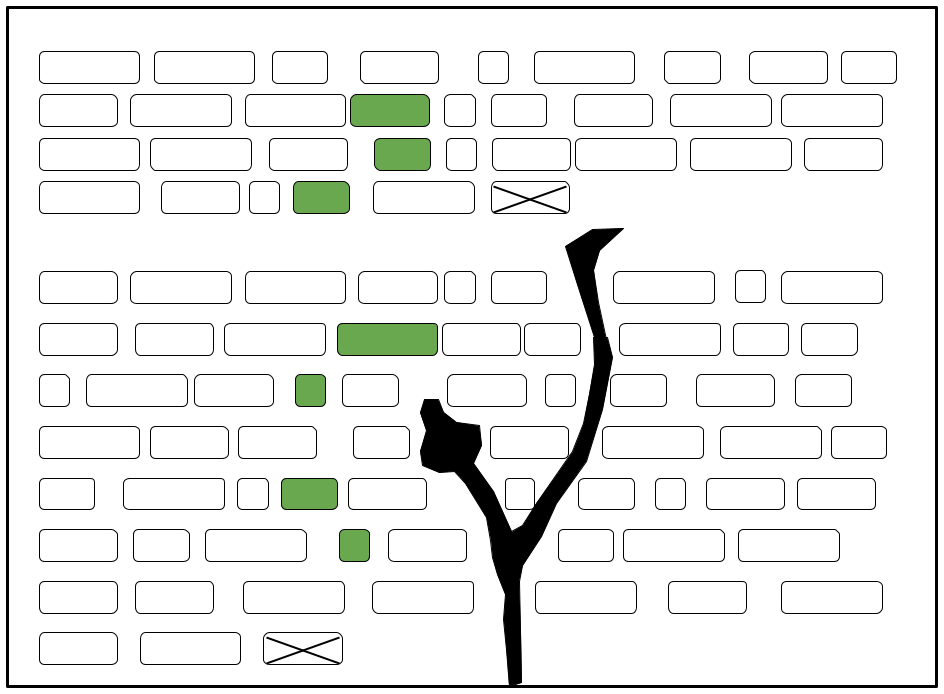}
          \caption{FOURTH}
          \label{fig:schem_fourth}
        \end{subfigure}
      \end{minipage}
    } 
    \caption{Schematic of Cohort Tokens. \textit{Each diagram shows a hypothetical folio page with two paragraphs of tokens, the second paragraph conforming around a drawing. The shaded tokens indicate cohort members. }}
    \label{fig:schematic_of_cohorts}
  \end{minipage}
\end{figure}

One may gain a more precise understanding of what tokens are represented by each of the positional cohorts by studying the Venn diagram in Figure \ref{fig:venn_diagram} and the schematic diagrams in Figure \ref{fig:schematic_of_cohorts}.  

\section{Analysis of Token Lengths}

Table \ref{tab:summary_of_cohorts} also includes the mean and standard deviation of the token lengths observed in each of the cohorts, where the length is defined by the glyph count within the token.\footnote{Multiple glyphs denoted as ligatures in the transliteration were counted individually.} We can see some distinct differences in these mean glyph counts. There are thirteen cohorts that each comprise a subset of the MIDDLE cohort and of these, ten have mean token lengths that are clustered within 2.5\% of the mean token length for the MIDDLE cohort.  The TOP and FIRST cohorts, however show outlier means that are 10\% and 12\% greater than that of the MIDDLE cohort respectively. Interestingly, the BEFORE cohort is also an outlier, showing a mean that is ~7.5\% lower.

To be clear, none of these mean token length differences between cohorts are very great. Furthermore, the variance of token length in each cohort is relatively broad. Therefore, little can be concluded from these measurements alone, but they are sufficient to indicate that there may be something more going on, and that a deeper look comparing the distributions of these token lengths is warranted.\footnote{Plots overlaying the probability mass distributions of token lengths for each pairs of cohorts can be found in the Supplemental Online Material.}

The data were therefore analyzed further by applying a statistical test of independence to the token length distributions of each pair of cohorts. 
A test of independence attempts to quantify the probability, or ‘$p$-value’, of a ‘null hypothesis’—that is, the hypothesis that the differences between two observed distributions can be attributed to chance alone.
The alternative hypothesis is that the two distributions are different, not by chance, but due to some underlying causal mechanism. 

Although it has been observed that the Voynichese token lengths follow binomial distributions very closely \cite{stolfi_vms_2000}, we have avoided assuming that this, or any other particular distribution, holds for every cohort.
Consistent with this, we have applied a $\chi^2$ (Chi-squared) test of independence, which makes no assumptions regarding the form of the probability distributions of the sampled populations being compared.  

The test does assume that the observations within each category of the distribution are independent. In our case, this would be the assumption that each value of token length within a distribution is independent. We would expect this to be the case since, given a  set of unique tokens employed by a scribe (presumably representing words in a language vocabulary), there is no reason to believe that the usage of tokens of any particular length would be related to the usage of tokens of any other length.

The $\chi^2$ test also depends on there being several observations within each value category; having an expected count of more than 5 in at least 80\% of the value categories is considered sufficient \cite{bewick_statistics_2003}.
This condition was met by all of our cohort distributions, but we have still combined the highest couple of value categories (glyph counts of 9 and 10) when needed to achieve at a minimum of 5 counts for each value of token length. 

Figure \ref{fig:chi2_matrix} presents the symmetrical matrix of $p$-values showing which pairs of cohorts pass the test. This, of course, depends on the $p$-value threshold chosen. A value of 0.05 is very common when applying statistical tests but we used a more stringent value of 0.01.  In the figure, the $p$-value is shown to five significant digits for each test pair. Darker cells indicate pairs of cohorts that pass the null hypothesis test.  This means that the two cohorts are similar enough that their differences may be explained by chance sampling. In other words, we cannot infer that they represent different underlying populations. Conversely, the lighter cells indicate a pair of cohorts governed by different causal phenomena, resulting in statistically different distributions of token length.

This matrix shows some interesting results. It indicates that the cohorts of MIDDLE, AFTER, SECOND, and FOURTH are all similar enough to each other that differences observed between them cannot be considered significant. It also indicates that the FIRST cohort is unique, showing a statistically significant difference from \textit{all} the other cohorts. We also see that the TOP cohort is different from most of the others although it passes the test against the AFTER cohort (meaning we simply cannot rule out that the differences are due to chance sampling). 
None of this is too surprising—differences in the tokens in the top line of paragraphs and in the first position of lines have been noted by many other Voynich researchers, although the observation has generally been regarding the greater presence of certain glyphs in these locations, not their token lengths.

What is notable from these tests, however, is the difference of the token lengths of the BEFORE and LAST cohorts from all other cohorts along with their similarity to each other.  We had anticipated, for reasons described earlier, that the token choices immediately prior to a drawing and to a lesser extent at the end of a line or script, may contain trace evidence of the scribe’s intentions. Nevertheless, we were surprised that these results, using formal tests of statistical significance, were so pronounced. 

Consider, for example, tokens appearing just before a drawing and those at the ends of lines. These two cohorts remain distinct from all others unless we reduce the $p$-value threshold below 0.005. This results in tokens at the ends of lines appearing similar to those immediately following a drawing. The $p$-value threshold must be lowered orders of magnitude further before one can attribute the differences between the tokens before a drawing or at the ends of lines (compared to any other cohort) to random chance alone.

In short, there is little ambiguity to be found in these results!

\customfig{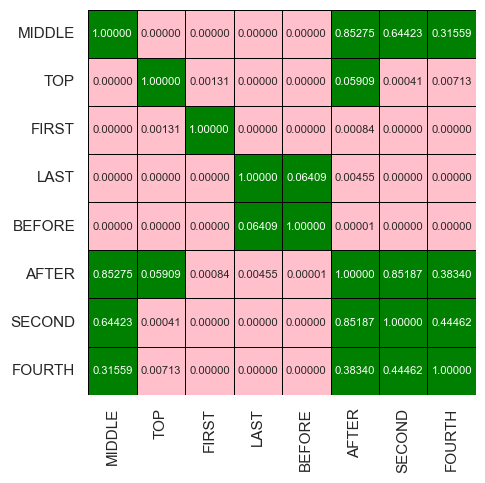}{$\chi^2$ Statistical Significance Matrix. \textit{Lighter cells indicate that the pair of cohorts that are most probably governed by different causal phenomena resulting in observed differences between their distributions of token lengths.}}{fig:chi2_matrix}{htbp}

Tests of significance using $p$-values are  based on solid statistical reasoning, and when properly applied and correctly interpreted, they provide the means to establish high confidence in analytical results. It should be noted, however, that the use of $p$-value as a measure of significance has raised some concerns among researchers in recent years. In fact, one  highly viewed article in Nature magazine dealt specifically with this concern \cite{nuzzo_scientific_2014},  and  the American Statistical Association subsequently issued a statement of caution in regard to the over-reliance on $p$-value testing \cite{wasserstein_asa_2016}.  

In the present analysis of token lengths, we selected a threshold that was well in excess of that commonly considered sufficient, and ensured that the statistical requirements for the validity of these tests were well met. But, given that the implications of the results are both subtle and potentially controversial,  it is worth repeating with precision what they do and do not tell us. 

\begin{mdframed}[backgroundcolor=gray!20, roundcorner=5pt]
The tokens found in certain positions on a page, relative to paragraph layouts and drawing intrusions, exhibit a distribution of token length that differs from that found in the larger population of tokens spread throughout the middle of the paragraphs. The statistical tests indicate that it is extremely improbable that these differences can be explained by random variations in sampling. This implies they must be due to some underlying causal mechanism that is related to the token’s position, although the analysis itself cannot reveal what that mechanism is.
\end{mdframed}

\section{Analysis of Token Positional Propensities}

So far, we have looked at differences in the distribution of token lengths, but we have not yet looked at which tokens might account for the differences. We now look at particular tokens regardless of their lengths, and how the propensity of particular tokens differs between cohorts. Are the shifts in mean token lengths between cohorts due simply to the scribe choosing tokens based on their length alone? Or are the shifts an incidental consequence of choosing particular tokens more often, or less often?

To explore those questions further, we have taken the middle locations within the paragraphs of Voynichese as representing the “base” population of tokens. The assumption is that, if the scribe is influenced at all by position when choosing tokens, then regardless of whether that choice is driven more by semantic content, a cipher process, or some language simulation device, the positional consideration will be least when writing tokens in these middle locations. And so the MIDDLE cohort becomes our reference cohort. We have then looked at each unique token across the lexicon of the study corpus, and compared its frequency of occurrence within each of the subject cohorts —TOP, FIRST, LAST, BEFORE, and AFTER—  to that within the MIDDLE cohort. We have then performed statistical significance tests to determine which tokens have a propensity to occur much more than expected (“affinitive”), or much less than expected (“aversive”), in each of the subject positions.   

In the previous analysis, a $p$-value threshold was chosen to determine if the token length distribution at specific page positions differed significantly from that of the reference cohort. We saw that the basic result was unchanged over a wide range of potential $p$-value thresholds. Here, however, we are assessing a large number of unique tokens to determine for each, whether it indicates a propensity to be used by the scribe in various specific positions. 

This ‘propensity’ should not be confused with ‘prevalence’. Measures of the occurrence counts and relative frequencies of tokens, have often been reported by Voynich researchers. While that can give a useful quantitative measure of token prevalence in particular positions, by itself it is reductively simplistic —and potentially misleading— for understanding usage propensity and whether the observed token usage is intentional due to behavioral or grammatical mechanisms, or an epiphenomenon due to sampling variations.

The better approach for our purpose is to consider the statistical significance and confidence level associated with a measurement of relative prevalence between positions before accepting it as an indication of propensity. We can do this by applying a binomial test, wherein we treat the occurrence of a particular token in the subject cohort as a binomial random variable whose probability is determined from the reference cohort. The binomial test, as an exact statistical procedure, does not presuppose the same conditions required by the $\chi^2$ test, notably the need for the number of observations of the token to be of sufficient size. 

We chose to use the same $p$-value threshold of 0.01, as used in the previous analysis, but to  get a sense of the sensitivity of the results to that choice, we first performed a parametric analysis. Figure \ref{fig:parametric_p_values} shows, for each cohort, how the number of unique tokens considered to have a significant propensity would vary if we changed the choice of $p$-value threshold.

\customfig{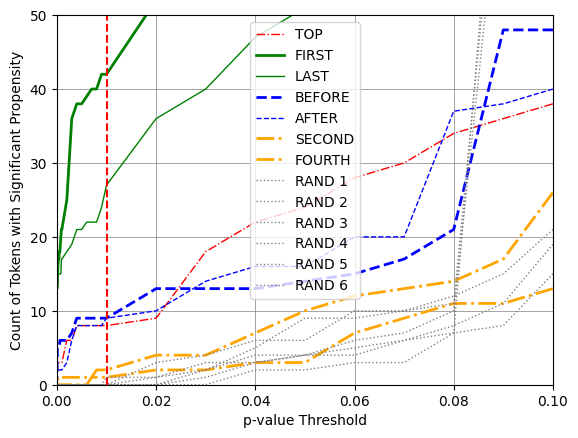}{Count of Tokens with Significant Propensity: Variation with $p$-value Threshold}{fig:parametric_p_values}{htbp}

We can see in the figure that as we raise the threshold, more and more of the unique tokens pass the test of significance. This is not unexpected since no matter how different the observed count of any particular token is from its expected count, selecting a high enough threshold will render it explainable as random sampling error.  But this trend also highlights the fact that the choosing $p$-value thresholds is always somewhat arbitrary. 

An additional insight from Figure \ref{fig:parametric_p_values} is notable. The figure shows the validation cohorts, including six random and two positional cohorts (SECOND and FOURTH). Although similar in size to the smaller subject cohorts and comprising hundreds of unique tokens (see Table \ref{tab:summary_of_cohorts}), they form a bundle of traces distinct from other traces in the plot, indicating few or no tokens with propensity at lower $p$-values. This reinforces the idea that the five subject cohorts significantly differ from the reference cohort in regards to usage of specific tokens with either higher (affinitive) or lower (aversive) propensity.

While the $p$-value tells us the probability of the observed data under the assumption of the null hypothesis (i.e. that the observed differences in the token counts in the subject and reference cohorts could happen by chance), it does not quantify the strength of evidence in favor of an alternative hypothesis (i.e. that the token’s usage in a particular position, being governed by a different underlying mechanism, occurs with a different probability than in the reference cohort).

For this reason, we have included a Bayesian measure of significance, the Bayes Factor ($B$), which is an alternative to traditional frequentist methods that rely on $p$-values. The Bayes Factor quantifies the strength of evidence in favor of one hypothesis over another, and is defined by the ratio of the observed data's likelihood under a candidate hypothesis to its likelihood under the null hypothesis.

In our case, this Bayes Factor represents how much more likely it is that the scribe’s token choices occur in the subject cohort according a different governing mechanism than that for his token choices in the reference (MIDDLE) cohort. The greater the Bayes Factor, the greater is our confidence in the hypothesis that the token truly has a highly affinitive, or highly aversive propensity to be used by the scribe in particular positions. 

Categorical interpretations for the Bayes Factor as a measure of evidence strength have been suggested elsewhere \cite{wei_review_2022} as follows:

\begin{tabular}{ll}
$B < 3$ & Worth only a bare mention \\
$3 \leq B < 20$ & Positive \\
$20 \leq B < 150$ & Strong \\
$B \geq 150$ & Very Strong \\
\\
\end{tabular}

While the $p$-value is bounded to the $[0, 1]$ range, the Bayes Factor is unlimited with a range of $[0,\infty]$.  So it is convenient to use the logarithm of $B$ instead.  The above values of 3, 20, and 150 happen to correspond closely to $log(B)$ values of 1, 3, and 5 respectively (where $log(…)$ designates the natural logarithm).

To gauge whether a particular token has a positional propensity, we chose the very conservative approach of using \textit{both} the binomial test and the Bayes Factor ‘Very Strong’ category. So for each cohort, we identified all tokens that resulted in both $p$-value$\le 0.01$ and $log(B)\ge 5$.

\begin{table}[H]
  \centering
  \includegraphics[width=\linewidth]{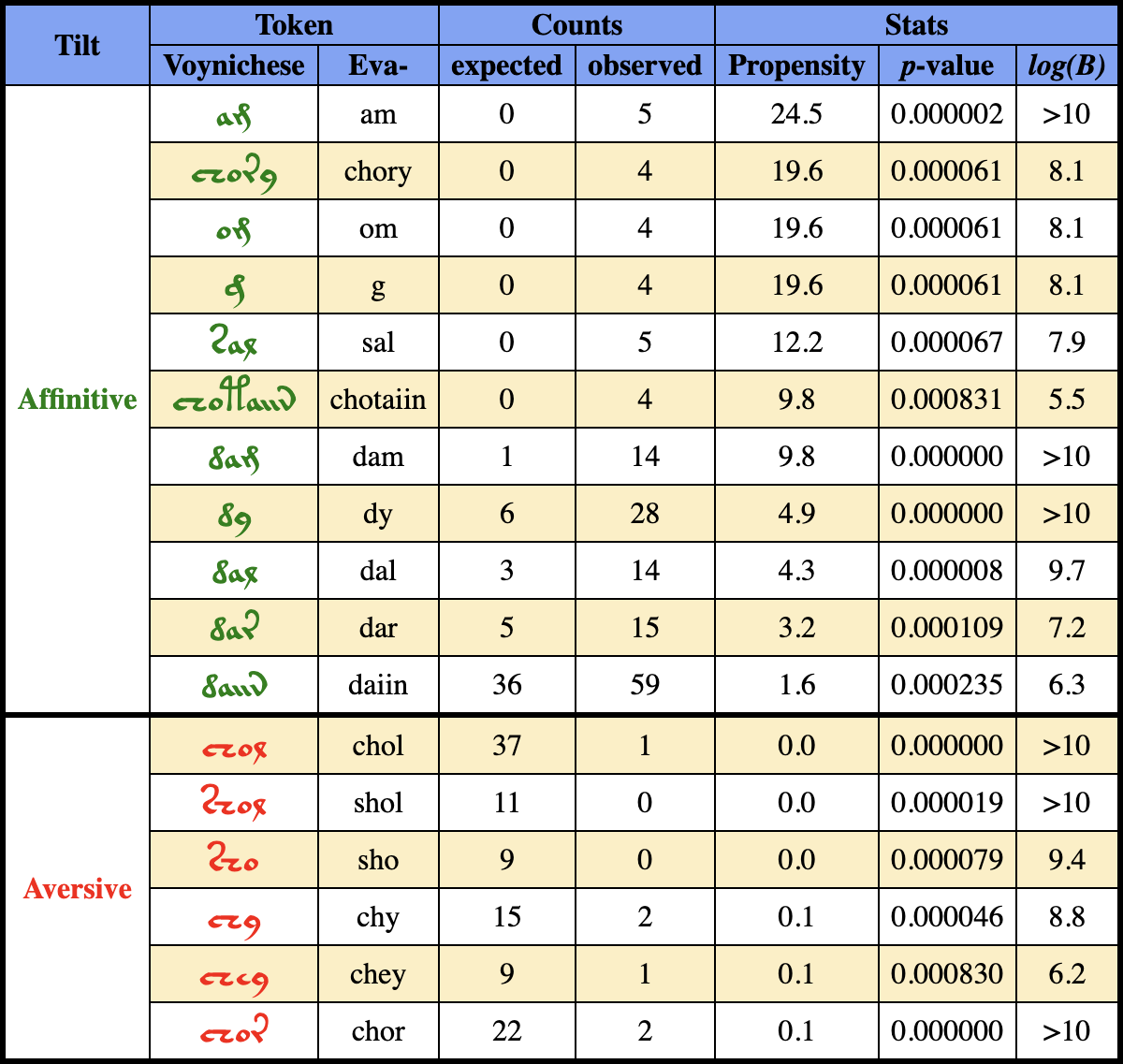}
  \caption{Tokens with Propensity for Last Position on a Line (Cohort: LAST)}
  \label{tab:token_propensity_last}
\end{table}

All five of the positions represented by the subject cohorts were found to have certain tokens showing statistically significant  propensity. This includes the positions that are intrinsic to the script itself (TOP, FIRST, and LAST), but surprisingly also the positions that are extrinsic in the sense that the position is defined by elements that are not part of the script (BEFORE and AFTER).  

All of the identified tokens have been tabulated by position and by their tilt, either affinitive or aversive. Due to space limitations, only two example tables are included here (Tables \ref{tab:token_propensity_last} and \ref{tab:token_propensity_before}).\footnote{All of our study results, along with additional analyses, discussions, and the full set of tables, are available in the Supplemental Online Material at https://www.quantumlynxresearch.com/research.}

Note that the additional Bayes Factor criteria caused rejection of some of the tokens otherwise counted for in the traces in Figure \ref{fig:parametric_p_values}, but ensured that the tables provide a catalog of only those tokens with very strong evidence of positional propensity that are also statistically significant.  

We have also listed in the tables a quantified measure of propensity defined as the ratio of the best estimates of the probabilities of the token occurring in the subject vs the reference cohorts.\footnote{We derive estimates of these probabilities directly from a simple ratio of observation counts; no additive smoothing is used as it would distort true zeros which are legitimate for this sparse data.}

\begin{table}[H]
  \centering
  \includegraphics[width=\linewidth]{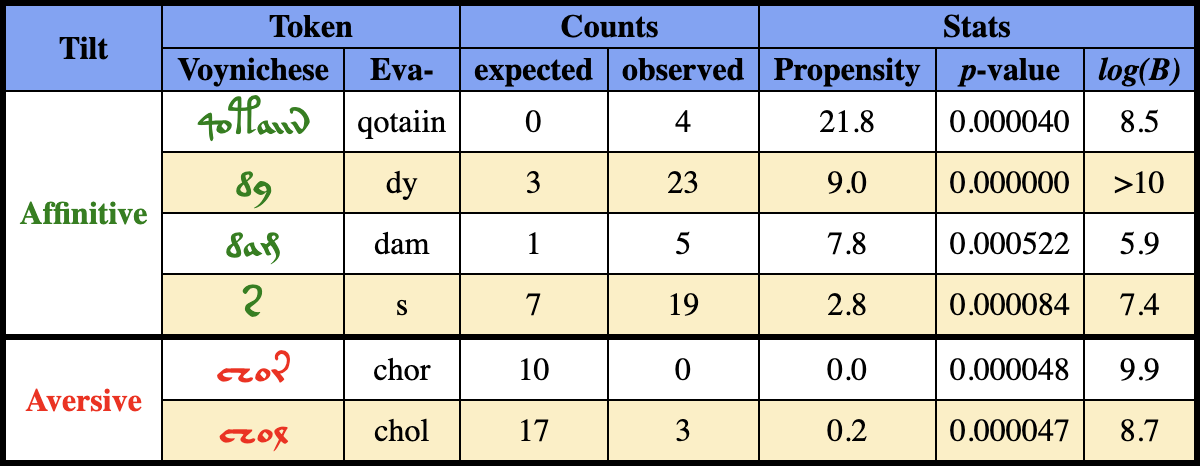}
  \caption{Tokens with Propensity for Position Immediately Before a Drawing Intrusion (Cohort: BEFORE)}
  \label{tab:token_propensity_before}
\end{table}

\section{Conclusions}

This study has produced several findings validated by formal tests of significance. They are not only significant in the formal statistical sense, but also in terms of their potential value toward understanding the nature and structure of the Voynichese script. 

The overarching conclusion from this study can be summarized as:

\vspace{\baselineskip}
\begin{minipage}{\linewidth}
\begin{mdframed}[backgroundcolor=gray!20, roundcorner=5pt]
The distributions of the unique Voynichese tokens found in the Voynich Manuscript depend not only on their position within paragraphs and lines of script (intrinsic positioning), but also on position in relation to the hard boundaries imposed by the presence of drawings (extrinsic positioning).
\end{mdframed}
\end{minipage}
\vspace{\baselineskip}

More specific conclusions include:
\begin{enumerate}[itemsep=0pt, topsep=0pt]    
\item Tokens immediately preceding the intrusion of a drawing on an otherwise continuous line of script, and tokens at the ends of lines, tend to be shorter than tokens located elsewhere.
    \item Tokens in the top lines of paragraphs, at the beginnings of lines, and immediately following a drawing intrusion, tend to be longer.
    \item Certain tokens have significant propensities to be either used or avoided by the scribe, depending on position. 
	\item Tokens exhibit positional propensity not only for positions dependent on the script itself, but for positions dependent on the drawings, which are extrinsic to the script.
\end{enumerate}

A catalog of all tokens with significant positional propensity has been compiled for use in further research.	

\section{Implications}

Of the several findings itemized in the previous section, the one regarding tokens adjacent to drawings is perhaps the most surprising.  Anomalies in the tokens and glyphs found in the top lines of paragraphs and at the beginnings and endings of paragraphs have been reported before, although few of those efforts have considered the statistical significance of the reported observations. We are not aware,  however, of any other efforts that have analyzed the Voynichese script in relation to  positions that are dependent on elements extrinsic to the script itself.  

The finding that there exists significant propensity for particular tokens immediately before and after drawings is not only unprecedented, but may also have radical implications for understanding the  nature of the manuscript, and particularly the Voynichese script.

A full discussion of potential implications is beyond the scope of this paper, but a couple of thought-provoking scenarios are worth mentioning.

\begin{table}[H]
  \centering
  \includegraphics[width=\linewidth]{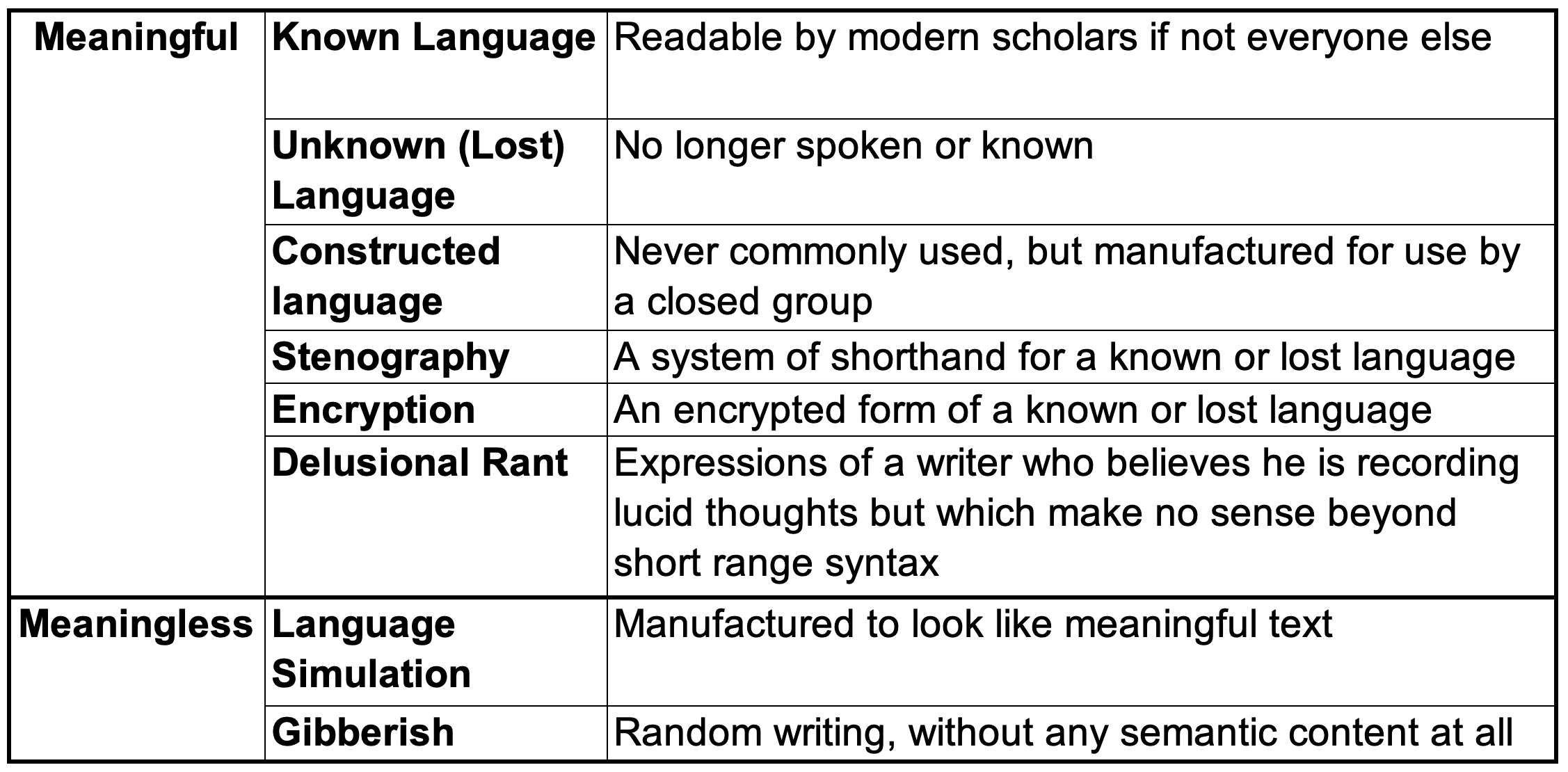}
  \caption{Possible Language Variants Underlying the Voynichese Script. \textit{This is a comprehensive list of possibilities; however, they are not all mutually exclusive.}}
  \label{tab:possibilities}
\end{table}\vspace{0.5\baselineskip}

\vspace{-0.5\baselineskip}
Table  \ref{tab:possibilities} summarizes possible language variants that could explain a script like Voynichese. Regardless of which of these is true, having tokens with a propensity related to their adjacency to drawings implies an unnatural coupling of elements that are not part of the script itself,  either to the semantic content of meaningful script or to the syntactic process that produced meaningless script.

For ‘meaningful’ script, the most plausible of the variants would seem to be selective use of stenography, where the scribe resorts to short-hand alternatives when choosing tokens next to a drawing. It is still difficult, however, to explain why the alternatives chosen would be biased to a small set of particular tokens unless perhaps they are a form of punctuation that is not tied directly to any prescribed semantic content. (And even this would not explain the 
affinitive \includegraphics[height=\baselineskip]{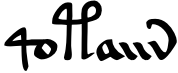}  and  aversive \includegraphics[height=\baselineskip]{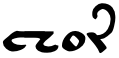} and \includegraphics[height=\baselineskip]{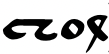}).

For ‘meaningless’ script, either of the variants  —a language simulation or gibberish— is conceivable, providing the scribe simply decided to make atypical choices for tokens when encountering a drawing, and did so with a preference to a favored list. 

In any case, our explanations for the observed extrinsic propensities are highly speculative at this stage, and so further research is underway.

Finally it is worth mentioning that, while we have applied considerable rigor in our analysis, we do not rule out the possibility of an overlooked systemic error or other plausible explanation for our results,  and we welcome review of the calculations, or suggestions regarding other interpretations of the results.


\bibliographystyle{histocrypt}
\bibliography{histocrypt}








\end{document}